\documentclass[
reprint,
superscriptaddress,
nofootinbib,
amsmath,
amssymb,
aps,
pra,
floatfix,
longbibliography,
showkeys,
]{revtex4-1}

\usepackage{amsmath}
\usepackage{graphicx}
\usepackage{lipsum}
\usepackage[toc,page]{appendix}

\usepackage{pgffor}
\usepackage{pdfpages}

\makeatletter
\AtBeginDocument{\let\LS@rot\@undefined}
\makeatother

\begin{document}
\title{Unexpected Benefits of Self-Modeling in Neural Systems}

\author{Vickram N. Premakumar}
\author{Michael Vaiana}
\author{Florin Pop}
\author{Judd Rosenblatt}
\author{Diogo Schwerz de Lucena}
        \affiliation{AE Studio, Venice, CA}
\author{Kirsten Ziman}
    \affiliation{Princeton Neuroscience Institute, Princeton University, Princeton NJ}
\author{Michael S. A. Graziano}
    \email[Corresponding author: ]{graziano@princeton.edu}
    \affiliation{Princeton Neuroscience Institute, Princeton University, Princeton NJ}
    \affiliation{Department of Psychology, Princeton University, Princeton, NJ}
\date{\today}

\begin{abstract}
Self-models have been a topic of great interest for decades in studies of human cognition and more recently in machine learning. Yet what benefits do self-models confer? Here we show that when artificial networks learn to predict their internal states as an auxiliary task, they change in a fundamental way. To better perform the self-model task, the network learns to make itself simpler, more regularized, more parameter-efficient, and therefore more amenable to being predictively modeled. To test the hypothesis of self-regularizing through self-modeling, we used a range of network architectures performing three classification tasks across two modalities. In all cases, adding self- modeling caused a significant reduction in network complexity. The reduction was observed in two ways. First, the distribution of weights was narrower when self-modeling was present. Second, a measure of network complexity, the real log canonical threshold (RLCT), was smaller when self-modeling was present. Not only were measures of complexity reduced, but the reduction became more pronounced as greater training weight was placed on the auxiliary task of self-modeling. These results strongly support the hypothesis that self-modeling is more than simply a network learning to predict itself. The learning has a restructuring effect, reducing complexity and increasing parameter efficiency. This self-regularization may help explain some of the benefits of self-models reported in recent machine learning literature, as well as the adaptive value of self-models to biological systems. In particular, these findings may shed light on the possible interaction between the ability to model oneself and the ability to be more easily modeled by others in a social or cooperative context.
\end{abstract}

\maketitle

\section*{Introduction}
Self-models have long been known to play a role in human cognition. The body schema, the brain’s model of the physical body, may be the first example of a predictive self-model to be extensively studied in the psychological literature \cite{graziano2002how, head1911sensory, holmes2004body}. More recently, other predictive self-models have been proposed, often described as forms of metacognition. For example, the brain may build a predictive model of its own process of decision making \cite{yeung2012metacognition}. A large body of work suggests that the brain also constructs an attention schema, or a predictive model of its own attentional processes \cite{graziano2011human, webb2015attention, webb2016effects, graziano2019rethinking, wilterson2020attention, mazor2023prospective}. The attention schema, in particular, has been studied in the context of social cognition \cite{kelly2014attributing, pesquita2016humans, guterstam2021temporo, guterstam2018implicit, guterstam2020other, ziman2023predicting}. Modeling one’s own attention and modeling the attention of others recruits overlapping systems in the brain, suggesting that learning one task contributes to the other \cite{kelly2014attributing, guterstam2020other}. Given this range of evidence, it appears that self-models serve an adaptive function in the human brain and may be related to the ability to construct models of other people in a social context.

To better understand how an attention schema might operate and what benefits or costs it might confer, several groups have studied artificial agents that contain attention schemas \cite{boogaard2017neurologically, wilterson2021attention, piefke2024computational, liu2023attention}. The addition of an attention schema can improve an artificial agent’s ability to control its own attention \cite{boogaard2017neurologically, wilterson2021attention, piefke2024computational}. The most pronounced effect of an attention schema, however, may be to improve performance in cooperative, multi-agent environments. A single agent, equipped with an attention mechanism and a predictive model of that attentional process, can show modest improvement on a solo task, but in an environment in which cooperation and interaction among multiple agents is necessary, the addition of an attention schema greatly improves performance \cite{liu2023attention}. The results from machine learning are therefore broadly consistent with the results from human
psychology. Self-models, and attention schemas in specific, confer a benefit in some contexts, with a possible emphasis on multi-agent or social cooperation. The underlying mechanism by which self-modeling improves performance is unknown.

Here we propose and test the hypothesis that when an agent self-models – when it is given the task of predicting its own internal states – a fundamental change occurs in the agent. In specific, we propose that the agent learns to reduce its own complexity, making its internal states simpler, more efficient, more predictable, and therefore more amenable to the task of self-prediction. In effect, a part of learning to self-model is learning to make oneself modelable. In this proposal, learning to self-model could be considered a type of network regularization.

This proposed principle of self-regularization through self-modeling could have broad implications, not just for machine learning, but also for understanding the evolution of biological systems and especially for understanding social cognition. In humans, theory of mind is the ability to intuit the internal states of other people and is fundamental to our social behavior \cite{baron1997mindblindness, frith2003development, wellman2018theory, wimmer1983beliefs}. Theory of mind only works, however, if the other person’s internal states are amenable to modeling. If self-modeling causes an agent to become more internally regularized and predictable, more amenable to being predictively modeled, then in principle it would cause the agent to become a better, more transparent target for theory of mind and a better member of a social, cooperative group. An agent with a less effective self-model would be less internally regularized, and therefore less amenable to being predictively modeled by self and others. One speculation is that social disabilities in people, such as autism spectrum disorder, trauma-related social difficulties, and some aspects of schizophrenia, may be partly related to incomplete self-models interfering with a person’s ability to resonate with others \cite{shalom2000developmental, cotraccia2021trauma, skorich2022integrated}. Even in simpler cases outside of human cognition, cooperation often depends on mutual predictability. Consider a pride of lions engaging in group hunting. One initiates the chase, several channel the flight path, and one ambushes the prey at a strategic location. These actions are coordinated and adjusted on the basis of continuous prediction of each other’s actions. In principle, learning to have more regularized internal states that are more predictable to everyone in the group should benefit cooperative interaction.

Given this broad theoretical context, it is important to test the underlying hypothesis that self- modeling causes self-regularization. Does self-modeling reduce complexity in neural networks?

We implemented self-modeling as an auxiliary regression task in a range of artificial neural networks. In addition to performing a primary task, the networks were responsible for predicting the values of subsets of their own hidden activations. The usefulness of auxiliary tasks in improving machine learning models has been studied from several perspectives \cite{caruana1997multitask, ruder2017overview, szegedy2015going}. Caruana’s work \cite{caruana1997multitask} on multitask learning highlighted the potential benefits of jointly learning multiple tasks to improve the performance of each individual task. Ruder \cite{ruder2017overview} provided an extensive overview of multi-task learning in deep neural networks, emphasizing the use of auxiliary loss to promote shared representations across tasks. The Inception architecture proposed by Szegedy and colleagues \cite{szegedy2015going} used auxiliary loss as a tool to avoid vanishing gradients by obtaining classification outputs from the model at several points in the architecture and including them in optimization alongside the model’s final output. In other contexts, self-supervised learning approaches have gained traction, leveraging pretext auxiliary tasks to learn useful representations without the need for human-labeled data. Misra and van der Maaten \cite{misra2020self} demonstrated the efficacy of a self- supervised auxiliary task carried out in advance. Similarly, Doersch and colleagues \cite{doersch2015unsupervised} proposed an unsupervised visual representation learning approach based on context prediction, showcasing the effectiveness of auxiliary tasks in learning informative representations.

Here we tested networks trained on three different classification tasks: the MNIST, the CIFAR- 10, and the IMDB tasks \cite{lecun1998gradient, krizhevsky2009learning, maas2011learning}. Our primary question was not whether the auxiliary task of self- modeling would improve performance on a classification task, but rather, whether it would decrease network complexity. We measured network complexity in two ways. First, we measured the width of the distribution of weights in the output layer, where narrower distributions implied reduced complexity. Second, we measured the local learning coefficient, or real log canonical threshold (RLCT), a local measure of complexity for a given function around its critical points \cite{watanabe2009algebraic}. We hypothesized that adding self-modeling as an auxiliary task to networks would reduce both measures of network complexity.

\section*{Methods}
\subsection*{Self-Modeling Mechanism}

\begin{figure}[h!]
    \centering
    \includegraphics[width=1\linewidth]{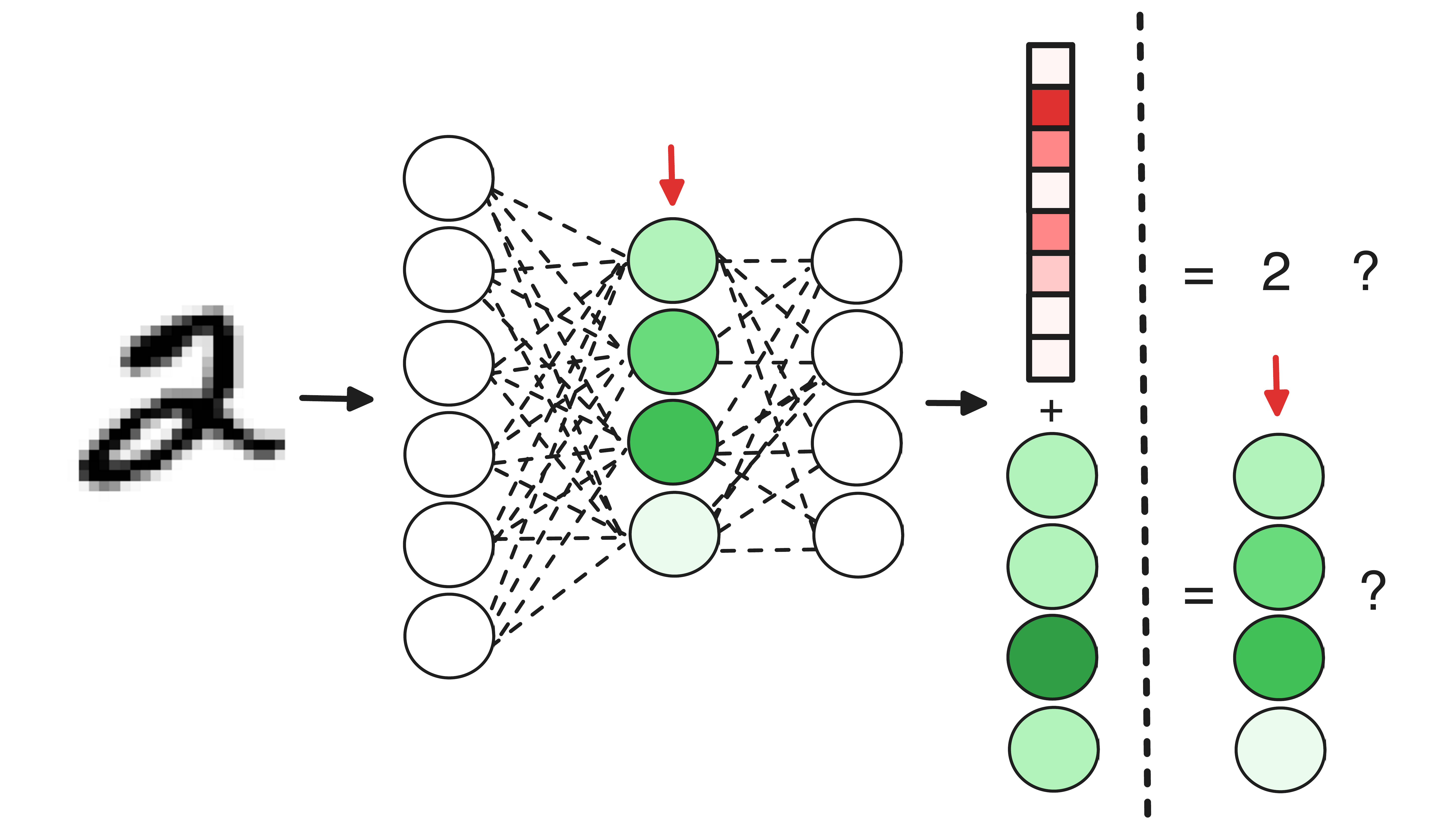}
    \caption{A schematic depiction of the self-modeling auxiliary task applied to an MNIST classifier. The red arrow indicates the selected layer to form the target of self-modeling. Evaluation is done by comparing classification outputs to the correct digit and the self-modeling outputs to the true activations of the network during its forward pass.}
    \label{fig:sm_cartoon}
\end{figure}

We trained several neural networks on classification tasks in the vision and language domains. As shown schematically in Figure \ref{fig:sm_cartoon}, to implement self-modeling, we augmented the output layer to predict both the class probabilities and the activations of specific target layers within the network. Therefore there were two additive loss terms, the classification loss for the classification task and the self-modeling loss, which was a regression term on intermediate activations. For the classification task, cross-entropy loss $L_c$ was used. The regression-style auxiliary task had loss $L_s$ where we selected specific layers $\{l_i\}$ of our architecture to form the target of self-modeling. During training, the $n$ activations of the neurons that composed these layers were retained and concatenated in a vector $a$.

The final layer of our architecture gave as output both logits for classification and regression predictions for the self-modeling task $\hat{a}$. The predictions and targets were passed to their respective loss functions and models were trained jointly on these two tasks. We included tunable weights $w_c, w_s$ as hyperparameters to change the relative importance of the main classification and self-modeling task. More explicitly, our loss function took the form

\begin{align*}
    \mathcal{L} &= w_c L_c + w_s L_s \\
     &= w_c L_c + \frac{w_s}{n} \left( \hat{a} - a \right) ^2 
\end{align*}

It's important to note that in this auxiliary loss, both $\hat{a}$ and $a$ were functions of the model weights. \textit{A priori}, the optimizer did not prefer adjusting the weights responsible for one or the other. During
training, the two could both be varied to find the best critical points under the joint loss function.

The inclusion of the extra piece of loss meant that we needed to be precise when comparing
learning dynamics and generalization performance between self-modeling variants and our
baseline networks. When we calculated complexity metrics we restricted
our models to their classification loss manifold.

\subsection*{Distribution of weights and RLCT measurement}

We employed several different neural architectures ranging in complexity from multi-layer perceptrons to residual networks. To probe model complexity and compare control implementations to self-modeling variants, we included two measures that provided insight on the development of model weights.

The first measure was the width in standard deviation of the distribution of weights in the final layer of the network. The magnitude, distribution, and exact values of weights have been used in myriad contexts to compare and select models \cite{goodfellow2016deep, hastie2009elements}. Here, we reported the spread of the distribution of weights in the final layer as a measure of model complexity. Two models which perform similarly can be differentiated by comparing the number of weights each one has that are far from zero. Many weights close to zero correspond to a node which can practically discard certain inputs, resulting in a smaller effective parameter count. Encouragement of smaller weight magnitudes is the goal of weight-norm regularization methods meant to improve the generalization ability of a trained model \cite{krogh1992simple, nowlan1992simplifying}.

We also included a more theoretically-founded measure of model complexity to help compare variants: the local learning coefficient or real log canonical threshold (RLCT) \cite{watanabe2009algebraic}. RLCT is a local measure of complexity for a given function around its critical points. Its initial use in statistical learning contexts was a measure of generalization meant to fill the role of parameter count in classical learning theory. We interpret the RLCT as effective model dimension and take the perspective that given two models with comparable performance, the one with a lower RLCT will be preferable. Lower RLCT indicates less overfitting and greater parameter efficiency. The quantity is difficult to calculate in practice and we followed Lau \cite{lau2023quantifying} to utilize stochastic gradient Langevin dynamics to estimate a localized variant of the RLCT. Unlike many other local measures of geometry, Furman \cite{furman2023estimating} showed it to be robust against loss rescaling, rebutting a common critique of such approaches \cite{dinh2017sharp}.

Before calculating either of these measures, we pruned the network of its self-modeling outputs in the final layer. In this way, the comparison was made between networks with identical structure which differed only in weights. This manipulation is important for the weight distribution metric and the RLCT, since both measure properties of learned critical points in weight space.

\section*{Results}

\subsection*{MNIST Classification}
We trained networks to perform the MNIST handwritten digit classification task \cite{lecun1998gradient}. The objective of the task was to predict the digit contained in a $28 \times 28$ grid of monochromatic pixels. Our variants consisted of several multi-layer perceptron networks with two layers, where we allowed the size of the hidden layer to vary over four values: $64, 128, 256,$ and $512$. We also allowed the weight of the self-modeling task to take five possible values: Auxiliary Weight (AW) $= 1, 5, 10, 20,$ and $50$. The target of the self-modeling was the hidden layer. Training occurred over $50$ epochs. Each model variant was trained 10 times, and we report mean and $95\%$ CI. The goal was to compare the results of the self-modeling networks to the baseline network with no self-modeling.

\begin{figure*}[t]
    \centering
    \includegraphics[width=\linewidth]{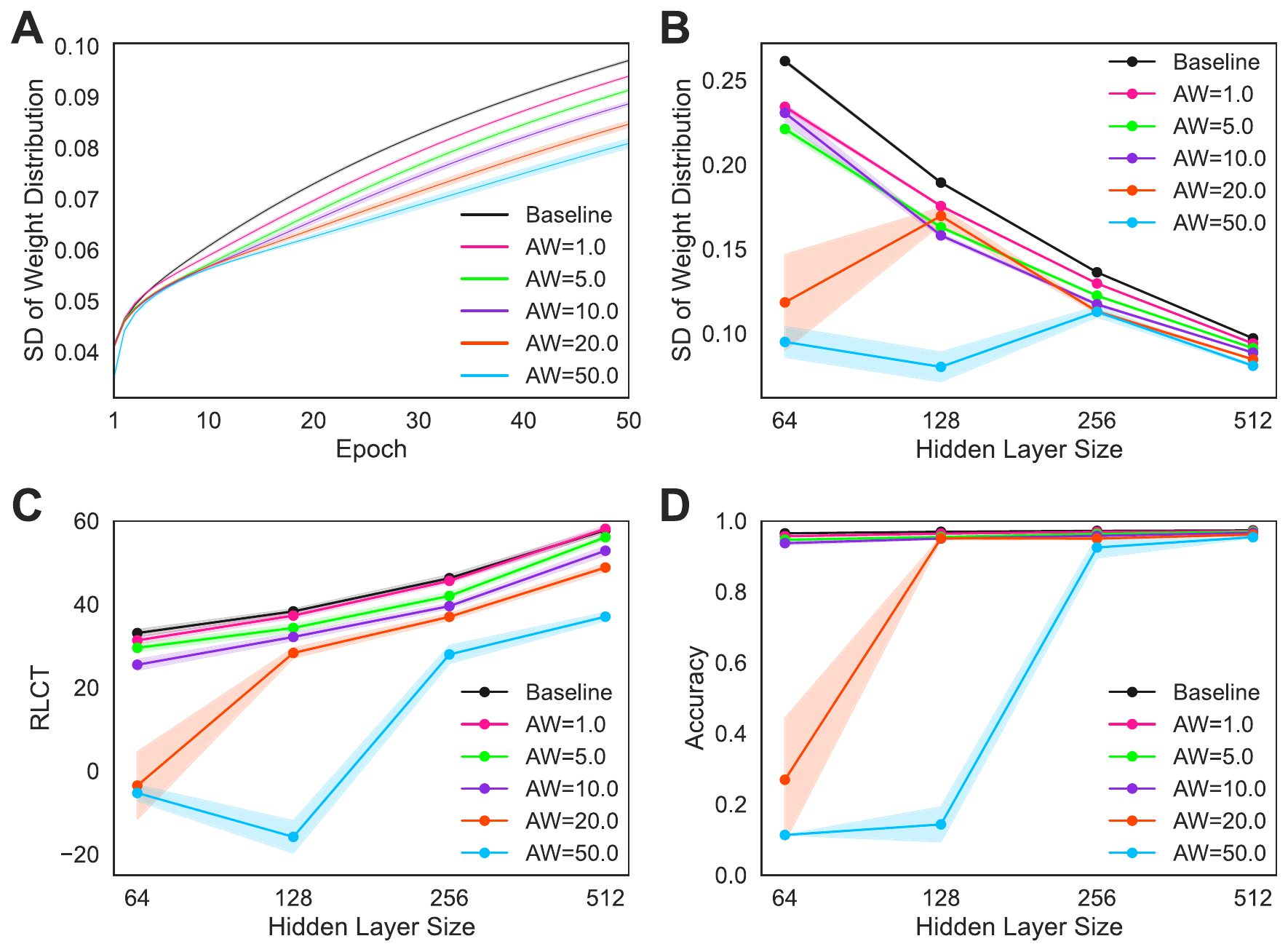}
    \caption{Results for networks performing the MNIST classification task. A. Y axis shows width of the distribution of weights in the final layer measured in standard deviation (SD). X axis shows training epoch ($1-50$). Data from baseline network with no self-modeling and from five networks that differ in self-modeling weights (AW $1-50$). Data are from the network architecture with a hidden layer size of $512$. Lines show mean of $10$ runs and $95\%$ CI. B. The width of the distribution of weights in the final layer (Y axis) as a function of the size of the hidden layer (X axis), for baseline network with no self-modeling and for networks with self-modeling at five weights. Data are from epoch $50$ of training. Points and error bars show means of $10$ runs and $95\%$ CI. C. The RLCT measure of network complexity (Y axis) as a function of the size of the hidden layer (X axis), for baseline network with no self-modeling and for networks with self- modeling at five weights. Data are from epoch $50$ of training. Lines show mean of $10$ runs and $95\%$ CI. D. Accuracy (\% correct) on the MNIST classification task (Y axis) as a function of the size of the hidden layer (X axis) for baseline network with no self-modeling and for networks with self-modeling at five weights. Data are from epoch $50$ of training. Lines show mean of $10$ runs and $95\%$ CI.}
    \label{fig:MNIST}
\end{figure*}

Figure \ref{fig:MNIST}A shows the results for the first measure of network complexity: the width (in standard deviation) of the distribution of weights in the final layer. A smaller width indicates more weights close to zero, and hence a sparser and potentially simpler network. The graph shows how the measure changed over training epochs, for networks with a hidden layer size of $512$, at five different weights for the self-modeling task. The measure of complexity was largest for the baseline network that had no self-modeling, and was reduced for networks equipped with self- modeling. Increasing the weight on the self-modeling task caused a systematic reduction in the width of the weight distribution. Self-modeling consistently decreased network complexity, at least by this measure.

Figure \ref{fig:MNIST}B shows the results for the final training epoch (epoch $50$), in networks with four different hidden layer sizes, for the baseline networks without self-modeling and for self- modeling networks with five different values of AW. Once again, the measure of complexity was reduced below baseline when the networks were equipped with self-modeling. In most cases, increasing the emphasis on the self-modeling task (increasing the AW) caused a reduction in the width of the weight distribution.

\begin{figure*}[t!]
    \centering
    \includegraphics[width=\linewidth]{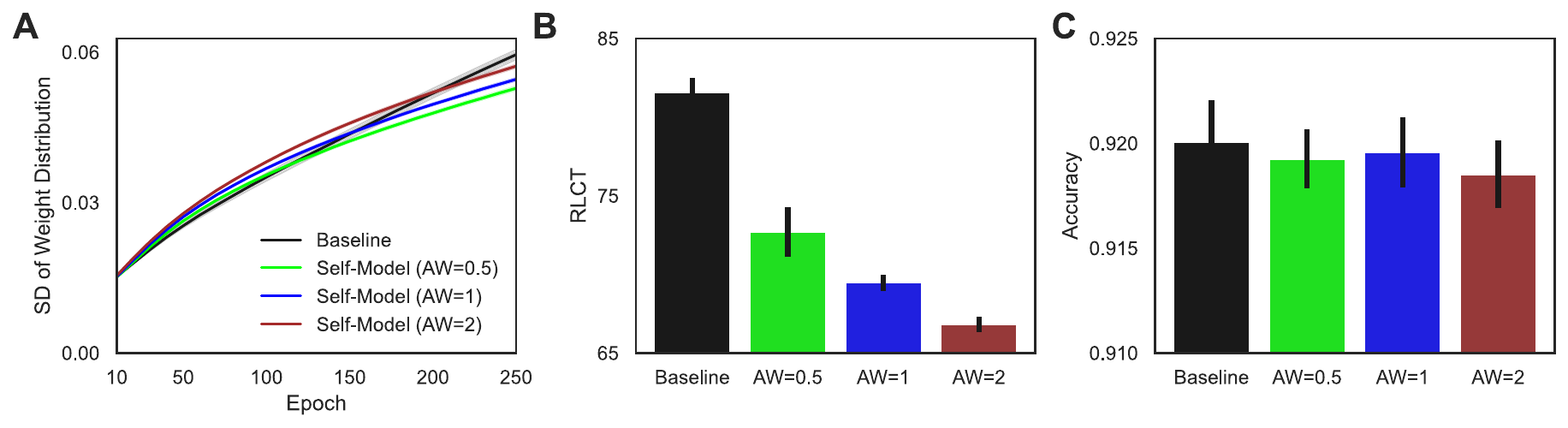}
    \caption{Results for networks performing the CIFAR-10 classification task. A. Y axis shows the width of the distribution of weights in the final layer measured in standard deviation (SD). X axis shows training epoch ($1-250$). Data from baseline network with no self-modeling and from networks with three different self-modeling weights (AW). Lines show mean of 10 runs and $95\%$ CI. B. The RLCT measure of network complexity for a baseline network and for three self- modeling networks that varied in AW. Data are from epoch $250$ of training. Bars show mean of $10$ runs and $95\%$ CI. C. Accuracy (\% correct) on the CIFAR-10 classification task for a baseline network and for three self-modeling networks that varied in AW. Data are from epoch $250$ of training. Bars show mean of $10$ runs and $95\%$ CI.}
    \label{fig:CIFAR}
\end{figure*}

Figure \ref{fig:MNIST}C shows that the results for our second measure of network complexity, the RLCT. Lower RLCT indicates less overfitting and greater parameter efficiency. Here the results are even more clear. The results are shown for the final epoch of training (epoch $50$), for four different hidden layer sizes, for the baseline networks without self-modeling and for self- modeling networks with five different values of AW. Networks that included self-modeling had lower RLCT than control networks without self-modeling. The effect was observed across all four network architectures. Moreover, the larger the emphasis on the self-modeling task (the larger the AW), the greater the reduction in the RLCT in every case. These results show that the addition of self-modeling systematically encouraged the optimization algorithm to find simpler solutions to learning the task. It is important to note the anomalous RLCT measures for the two smallest networks (hidden layer sizes of $64$ and $128$), trained at the two largest self-modeling weights (AW of $20$ and $50$). We suggest that in these specific cases, the self-modeling task weight was too great for the network, resulting in poor ability to learn the primary task. As a result, the RLCT could not be calculated around a critical point, because no such critical point was found by the network, resulting in anomalous values of the RLCT.

Figure \ref{fig:MNIST}D shows the accuracy of the network on its primary classification task, in the final epoch of training (epoch $50$), for the baseline networks without self-modeling and for self-modeling networks at the five different values of AW. This metric is calculated only on the held-out test
dataset ($N=10,000$). The baseline performance was close to $100\%$. For most self-modeling variants of the network, accuracy was only marginally affected. For the smallest networks (hidden layer size $64$ and $128$), when the self-modeling weight was at the higher values (AW of $20$ and $50$), the network suffered a breakdown in learning the primary task, showing low mean performance, presumably because the weight placed on the auxiliary task was great enough to interfere with the primary task. This result underscores the importance of finding a self-modeling training weight in a good operating range for the particular network architecture, that does not break the performance of the agent. That operating range is likely to be different for different network architectures.

At least by the two measures used here (the width of the distribution of weights and the RLCT metric), adding self-modeling as an auxiliary task reduced network complexity. The finding was particularly clear for the more theoretically-founded measure of complexity, the RLCT \cite{watanabe2009algebraic}. The result was consistent over a range of hidden layer sizes and auxiliary task weights. Self-modeling did not improve performance accuracy on the primary task. Instead, the key finding here is that, based on the complexity metrics, the networks became more efficient. The effect could be viewed as a form of regularization. The additional training incentive of predicting internal state caused the system to choose simpler functions to accomplish the main classification task.

\subsection*{CIFAR-10 Classification}

\begin{figure*}[t]
    \centering
    \includegraphics[width=\linewidth]{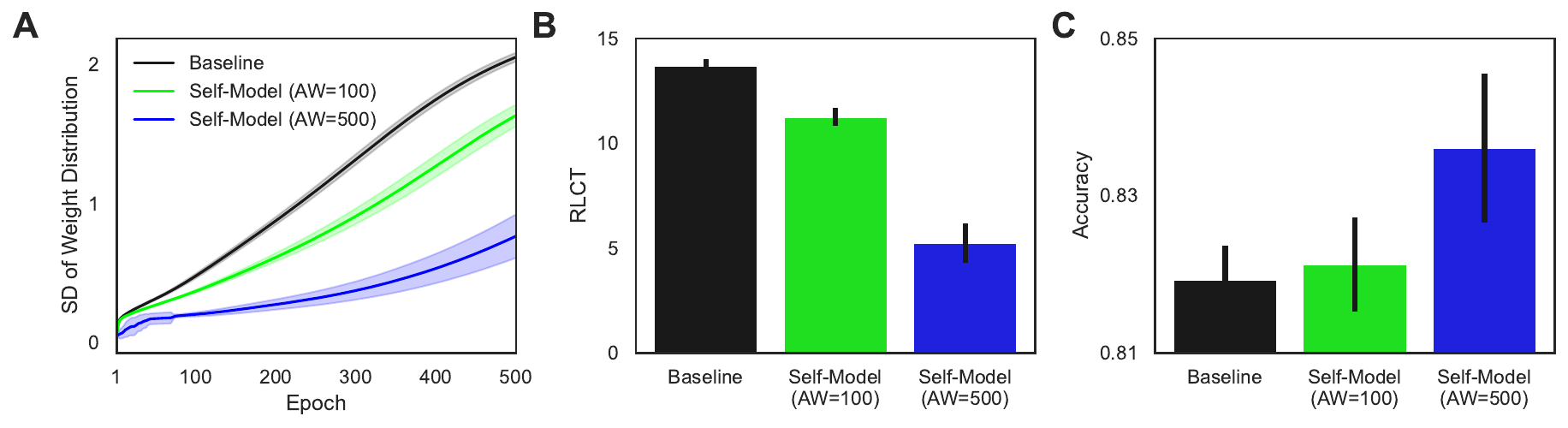}
    \caption{Results for networks performing the IMDB classification task. A. Y axis shows the width of the distribution of weights in the final layer measured in standard deviation (SD). X axis shows training epoch ($1-500$). Data from baseline network with no self-modeling and from networks with two different self-modeling weights (AW). Lines show mean of $10$ runs and $95\%$ CI. B. The RLCT measure for the baseline network, a self-modeling network with AW $= 100$, and a self-modeling network with AW = $500$. Data are from epoch $250$ of training. Bars show mean of $10$ runs and $95\%$ CI. C. Accuracy on the IMDB classification task for the baseline network, a self-modeling network with AW $= 100$, and a self-modeling network with AW $= 500$. Data are from epoch $500$ of training. Bars show mean of $10$ runs and $95\%$ CI.}
    \label{fig:IMDB}
\end{figure*}

After observing the simplifying effects across several MLP variants in the MNIST task, we
selected a task that demanded a more sophisticated architecture to test the generality of the
results. We used the ResNet18 architecture to perform classification on the CIFAR-10 dataset \cite{he2016deep, krizhevsky2009learning}. In this task, low-resolution color images of objects are classified into ten categories (e.g. automobile, bird, cat, and so on). The ResNet architecture is essentially a sophisticated feature extractor, identifying macroscopic visual features and passing them to downstream layers which are responsible for the final classification decision. We modified the standard ResNet architecture by adding a single linear hidden layer between the final convolutional block and the output layer. The target of the self-modeling encompassed the hidden layer and the input to the hidden layer (this input included the output of the fourth convolutional block and the skip-connections from the third block to the hidden layer). In these ways the architecture was different from that described above for the MNIST task. We found that in this ResNet architecture, smaller values of AW were sufficient to drive changes in the self- modeling versions, and therefore we used AW values of $0.5$, $1$, and $2$. Because the CIFAR-10 task is more difficult and requires more training than the MNIST task, we trained it on $150$ epochs.

Figure \ref{fig:CIFAR}A shows the results for the first measure of complexity, the width of the distribution of weights in the final layer of the network. The graph includes results from a baseline network without self-modeling and from three self-modeling networks that varied in their auxiliary task weights (AW$= 0.5, 1,$ and $2$). By the final training epoch, a pattern emerged: the weight distributions were widest for the baseline network indicating greatest network complexity, and were reduced for networks that contained self-modeling. By this measure, adding self-modeling caused a reduction in network complexity. Unlike in the MNIST example, in the present case increasing the emphasis on self-modeling (increasing AW) did not cause a systematic decrease in the width of the weight distribution.

Figure \ref{fig:CIFAR}B shows the results for the RLCT measure in the final epoch of training (epoch $250$), for the baseline networks without self-modeling and for self-modeling networks with three different values of AW. Once again, the baseline network, without a self-model, had the highest RLCT.
Networks that included self-modeling had lower RLCT scores. The effect was particularly clear in this case. The greater the emphasis on the self-modeling task (the larger the AW), the lower the RLCT score.

Figure \ref{fig:CIFAR}C shows the accuracy on the held-out test dataset for the classification task ($N=10,000$). Once again, self-modeling did not improve performance accuracy on the primary task. Instead, the key finding is that the additional training incentive of predicting internal state caused the networks to choose simpler functions to accomplish their tasks. Self-modeling reduced network complexity, acting as a form of self- regularization.

\subsection*{IMDB Classification}

Finally, we tested a third type of network trained on the IMDB dataset \cite{maas2011learning}. The task was to categorize the sentiment of a movie review presented as a block of text. We used a network architecture that consisted of a simple embedding layer connected to a linear hidden layer, which was connected to a linear output layer. The self-modeling was targeted at the hidden layer, the AW was set to either $100$ or $500$, and training extended to $500$ epochs.

Figure \ref{fig:IMDB}A shows the effect of self-modeling on the width of the distribution of weights in the final layer. The baseline network, without any self-modeling, had consistently wider weight distributions throughout training. The network that included self-modeling at a weight of $100$ showed narrower weight distributions than the baseline, and the network that included self- modeling at a weight of $500$ showed even narrower weight distributions.

Figure \ref{fig:IMDB}B shows a similar effect for the RLCT score. The RLCT was highest for the baseline network without a self-model, was reduced when self-modeling was present at a weight of $100$, and was further reduced when self-modeling was present at a weight of $500$. Figure \ref{fig:IMDB}C shows the accuracy of the network in performing the classification task on the held-out test dataset ($N=25,000$). Self-modeling slightly improved performance especially for the network with the larger self-modeling weight of $500$. These results show that in yet another set of networks, this time performing a textual task instead of an image-based task, self-modeling once again reduced network complexity according to both measures.

\section*{Discussion}
\subsection*{Effect of self-modeling on network complexity}

Self-models have been a topic of great interest for decades in studies of human cognition and more recently in machine learning. Here we hypothesized that learning a self-model should change networks in a fundamental way. In this hypothesis, when a neural network learns to predictively model its own internal states, to better perform that task, it learns to make itself internally simpler, more regularized and more easily predictable. This self-regularization through self-modeling may help to explain some of the benefits that have been seen in artificial networks that incorporate self-models as an auxiliary task, and it may also help to explain the adaptive value of self-models in biological brains.

In the present study, we tested the hypothesis of self-regularizing through self-modeling. We used a range of networks with different architectures that performed three different classification tasks (MNIST, CIFAR-10, and IMDB classifications), and we varied parameters including layer width and auxiliary task training weight. In all cases, we found that adding a self-modeling mechanism caused a reduction in network complexity. This reduction was observed in two ways. First, the distribution of weights was narrower when self-modeling was present. Encouragement of weight magnitudes that are clustered more tightly around smaller values is a goal of weight- norm regularization methods meant to improve the generalization ability of a trained model \cite{krogh1992simple, nowlan1992simplifying}. The second measure of network complexity that we used, the RLCT, was also consistently smaller when self-modeling was present. RLCT is a local measure of complexity for a given
function around its critical points \cite{watanabe2009algebraic}. Lower RLCT indicates less capacity for overfitting and greater parameter efficiency. Not only were both measures of complexity reduced for networks that included self- modeling, but in most cases the reduction became more pronounced as greater emphasis was placed on the auxiliary task of self-modeling. The RLCT measure showed this effect particularly clearly. These results strongly support the hypothesis that self-modeling is more than simply a network learning to predict itself. Self-modeling has a fundamental, restructuring effect on networks, reducing complexity and increasing parameter efficiency.

We did not predict any specific improvement in performance on the primary, classification task. In most cases, adding self-modeling either had no effect or slightly reduced accuracy. Yet it did sometimes improve accuracy in the primary task. On the IMDB classification tasks, small
improvements in performance were obtained. At least from the perspective of strict parameter count, one might expect that the self-modeling auxiliary task, by adding outputs to the final
layer, would always reduce accuracy on the primary task. The fact that adding self-modeling can
in some cases improve accuracy may seem counterintuitive. It might perhaps be partly explained
by the beneficial effects of reduced complexity and greater network efficiency in the presence of
the self-modeling task. The result is consistent with previous findings in which adding an
attention schema, a specific form of self-modeling, improved task performance for some tasks \cite{boogaard2017neurologically, wilterson2021attention, piefke2024computational, liu2023attention}.

It is important to remember the relative simplicity of the learning tasks used here. Image and sentence classification do not offer a tremendously rich set of patterns for the network to learn when compared to the capabilities of many modern systems, but they can test the hypotheses in systems where measures of complexity can be more easily computed, and they provide an initial step in understanding the fundamental principles of self-modeling. We suggest that the same principles may apply to machine learning at greater levels of complexity and to biological systems.

\subsection*{Possible Cooperative benefits of self-modeling}

We propose that networks that learn to self-model are not only better at predictively modeling themselves, but are also trained in such a way that they should be better at being predictively modelled by other agents. They have optimized themselves to be more modelable. They should be better members of an ensemble where self-modeling and mutual-modeling are advantageous. Like two dancers that have an exquisite ability to intuit each other’s moves, or lions that hunt in cooperative packs, or simpatico people who “get” each other because they have good theory-of- mind models of each other’s internal states, agents that self-model may restructure themselves to be better at predicting themselves, predicting others, and being predicted by others. It could even be that the evolution of predictive self-models in individual animals allowed for the eventual evolution of ensembles of animals that engage in mutually-predictive, complex patterns of social cooperation.

\section*{Acknowledgements and Funding Sources}

This research was supported by AE Studio grant 24400-B1459-FA010.

\medskip

\bibliographystyle{apsrev4-1}
\bibliography{refs}

\clearpage



\section*{Supplementary material}

# Supplemental Information for Unexpected Benefits of Self-Modeling in Neural Systems

Here we provide additional technical detail on the experiments described in the main text. The goal is to sufficiently describe our protocol for the sake of reproducibility and transparency. In all cases, the experiments were conducted across 10 random seeds and means with 95% CI reported. The code requried to run these experiments can be provided upon request.

## MNIST

The MNIST classifier we constructed is a simple multi-layer perceptron with one hidden layer. The width in neurons of the hidden layer were varied among $[512, 256, 128, 64]$. The raw image data was mean-normalized before being passed to the network. Additional details of the hyperparameters are given in Table 1.

## CIFAR-10

The classifier we constructed for testing self-modeling in the CIFAR-10 dataset is slightly more complex. We used the ResNet-18 architecture with an additional hidden layer of width 2000 between the final residual block and the output layer. Again, raw image data was mean-normalized in these experiments. Additional details of the training hyperparameters are given in Table 1.

We also elected to utilize data augmentation as a regularization method in this experiment. The case for this was mainly for replicating published performance benchmarks on the dataset, and we found that the self-modeling regularization effects persisted with this inclusion. Data augmentation was achieved by applying randomized affine transformations with rotations bounded by $\pm 15°$ and translations in a range of a tenth of the raw image dimension. We also allowed random horizontal flips of the images.

## IMDB

The text classification task utilized a custom model to conduct a series of text classification tasks on the IMDB sentiment dataset. The text embedding was generated using the EmbeddingBag layer in PyTorch. Specifically, the text classification model is defined with an embedding dimension of 64, a vocabulary size of $100,683$, and hidden layer size of 128. Hyperparameters of the optimization procedure are given in Table 1.

|  | **MNIST** | **IMDB** | **CIFAR** |
| --- | --- | --- | --- |
| Batch Size | 64 | 64 | 64 |
| Optimizer | SGD | SGD | Adam |
| Learning Rate | 0.005 | 0.1 | 0.001 |
| Momentum | 0.9 | 0.0 | n/a |
| Nesterov | True | False | n/a |

Table 1: Hyperparameters for MNIST, IMDB, and CIFAR Experiments

## RLCT Calibration

Across the three datasets, we utilized the real log canonical threshold, or learning coefficient, to measure effective model complexity [1]. Analytical calculation of RLCT is impossible in almost all cases of interest for deep learning. In practice, the measurement relies on the use of a stochastic sampling near critical points in the loss function. As in the use of stochastic optimization in training models, there are few explicit quantitative indications of optimality. Hence, we rely on heuristics to determine the right parameters to approximate the quantity. We use the Stochastic Gradient Langevin Descent (SGLD) method to do the estimation. This approach requires hyperparameter tuning that is done independently of the model training protocol. The two parameters we varied were the learning rate, or step size, of updates in weight space and the localization parameter which controls the strength of a quadratic focusing effect meant to keep moves inside a neighborhood of the trained model's weights. To summarize our choices, the MNIST and the CIFAR-10 experiments used lr = 0.0001 and localization = 1000, and the IMDB experiment used lr = 0.001 and localization = 10.

Our approach involved a grid search over many different values of the learning rate and localization. We selected the parameters for the final calculation using a qualitative appraisal of the loss curves. We should see a smooth increase to a plateau in loss. One example of this from our ResNet experiments is shown in 1. Rapid increases may indicate parameters that do not sample the neighborhood of the critical point sufficiently. A decrease in loss represents continued learning, and especially in the cases of our over-weighted self-modeling variants, shows that the point in weight space identified by training is not a loss minimum.

We additionally report the MALA acceptance rate at the end of the sampling process. Although with SGLD, we expect this to be high ($\sim 1$), the parameters we have chosen seem to have relatively low MALA acceptance rates. We believe this is likely due to the omission of learning rate scheduling from the sampling procedure. When the learning rate does not decrease, the algorithm imitates gradient descent and never transitions to the Langevin dynamics regime, so we should not expect the algorithm to reach a point where the acceptance rate is high. It is not clear to us whether or not including scheduling would actually improve our estimates, but it does make the MALA rate relatively less indicative of success.

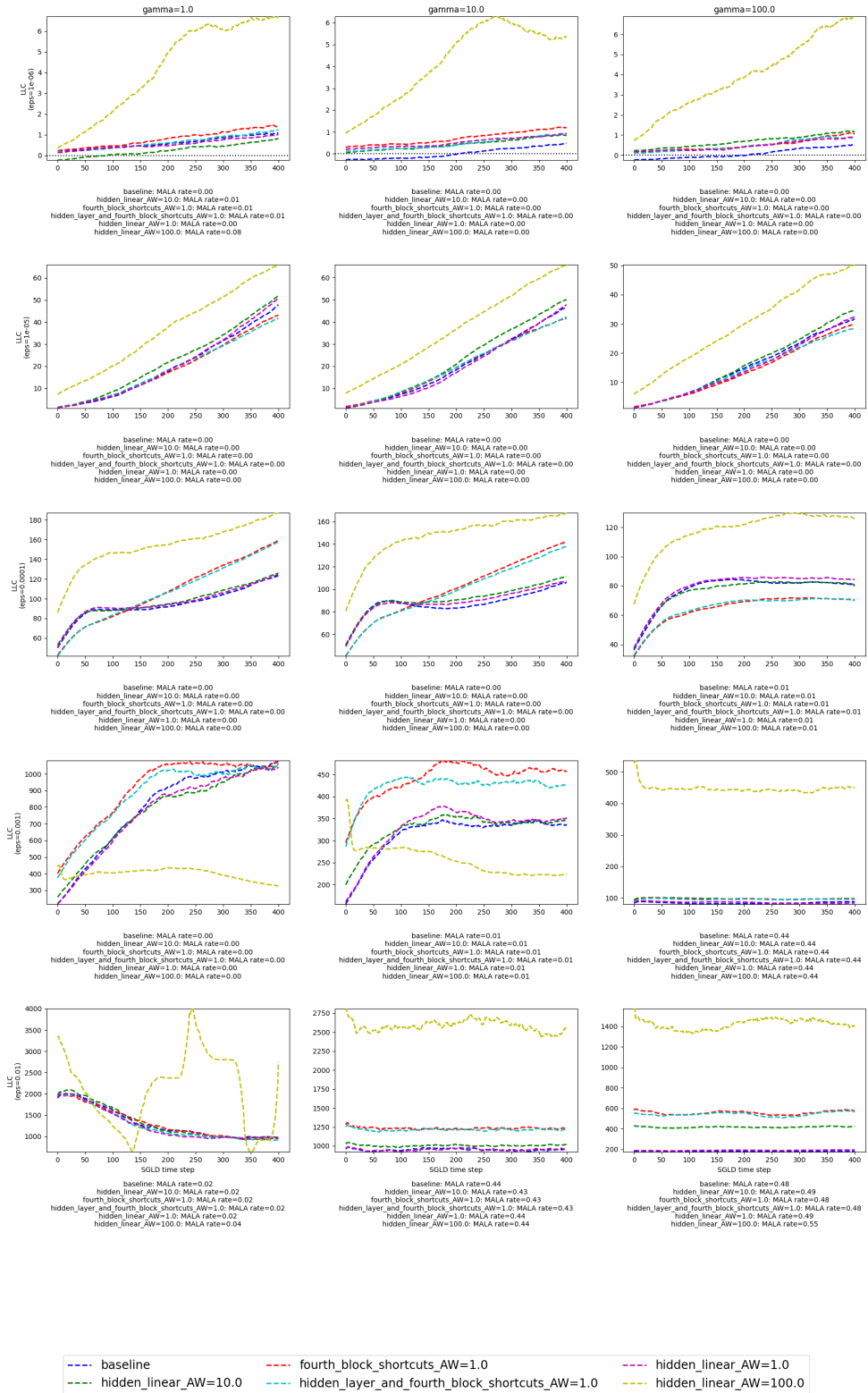

Figure 1: Hyperparamter optimization for the ResNet18 RLCT estimation process using SGLD. We applied the process to several of our trained experimental variants and found the optimal parameters at learning rate = 0.0001 and localization = 100.



\end{document}